%% file: advanced_robotics_ewerton_2019.tex
\documentclass[]{interact}

\usepackage{mathtools}
\usepackage{amsmath}
\usepackage{bbold}
\DeclarePairedDelimiter\ceil{\lceil}{\rceil}
\DeclarePairedDelimiter\floor{\lfloor}{\rfloor}

\usepackage{epstopdf}
\usepackage[caption=false]{subfig}

\usepackage[numbers,sort&compress]{natbib}
\bibpunct[, ]{[}{]}{,}{n}{,}{,}
\renewcommand\bibfont{\fontsize{10}{12}\selectfont}
\makeatletter
\def\NAT@def@citea{\def\@citea{\NAT@separator}}
\makeatother

\theoremstyle{plain}

\theoremstyle{definition}

\theoremstyle{remark}

\usepackage{wrapfig}

\begin{document}

\articletype{FULL PAPER}

\title{Assisted Teleoperation in Changing Environments with a Mixture of Virtual Guides}

\author{
\name{Marco Ewerton\textsuperscript{a, b}\thanks{CONTACT Marco Ewerton. Email: marco.ewerton@idiap.ch}, Oleg Arenz\textsuperscript{a}, and Jan Peters\textsuperscript{a, c}}
\affil{\textsuperscript{a}Intelligent Autonomous Systems Group, Department of Computer Science, Technische Universit\"at Darmstadt, Hochschulstr. 10, 64289 Darmstadt, Germany; \textsuperscript{b} Perception and Activity Understanding Group / Robot Learning and Interaction Group, Idiap Research Institute, Rue Marconi 19, 1920 Martigny, Switzerland; \textsuperscript{c}Max Planck Institute for Intelligent Systems, Max-Planck-Ring 4, 72076 T\"ubingen, Germany}
}

\maketitle

\begin{abstract}
\input{abstract.tex}
\end{abstract}

\begin{keywords}
teleoperation; policy search; variational inference; movement primitives; Gaussian mixture models
\end{keywords}

\section{Introduction}
\label{sec:intro}
\input{intro.tex}

\section{Connection to Prior Work} 
\label{sec:connection_to_prior_work}
\input{connection_to_prior_work.tex}

\section{Learning a Mixture of ProMPs} 
\label{sec:learning_promps_through_VIPS}

\input{learning_ProMPs_through_VIPS.tex}

\section{Computing the Haptic Cues}
\label{sec:haptic_feedback}
\input{haptic_feedback}

\section{Adapting the Mixture Weights by Updating the Belief}
\label{sec:belief_update}
\input{belief_update.tex}

\section{Adapting the Plans Online}
\label{sec:online_adaption}
\input{online_adaption.tex}

\section{Experiments}
\label{sec:experiments}
\input{experiments.tex}

\section{Conclusion and Future Work}
\label{sec:conclusion}
\input{conclusion.tex}

\section*{Funding}

The research leading to these results has received funding from the European Union's Horizon 2020 research and innovation program under grant agreement No. 640554 (SKILLS4ROBOTS).

\bibliographystyle{tfnlm}
\bibliography{bibliography_database}

\appendix

\section{Definition of the Shifting Operator}
\label{sec:definition_of_the_shifting_operator}
The shifting operator $\mathcal{T}_{\Delta \nu}(p)$ is defined as a function that maps a discrete probability distribution $p$ over $T_o$ phases to a new distribution over phases by shifting its probability mass by $\Delta \nu$ into the future (see Section~\ref{sec:updating_the_belief_about_the_phase}). Formally, $\mathcal{T}_{\Delta \nu}$ is a function 
\begin{equation}
\mathcal{T}_{\Delta \nu}: \mathcal{S}^{T_o-1} \mapsto \mathcal{S}^{T_o-1},
\label{eq:T_delta_nu}
\end{equation}
where $\mathcal{S}^n$ denotes the n-simplex. The $i$th vertex of the shifted simplex $\mathcal{T}_{\Delta \nu}(p)^{(i)}$ is defined as
\begin{equation} 
\mathcal{T}_{\Delta \nu}(p)^{(i)} = 
     \begin{cases}
     \begin{multlined}
       \mathbb{1}_{i - \floor*{\Delta \nu} > 1} \left[ (1 - \mathbb{1}_{\Delta \nu < 1} [(\Delta \nu - \floor*{\Delta \nu})]) p^{(i - \floor*{\Delta \nu})}  \right]  \\
       + \mathbb{1}_{i - \ceil*{\Delta \nu} > 1} \left[ (\Delta \nu - \floor*{\Delta \nu}) p^{(i - \ceil*{\Delta \nu})} \right]  
     \end{multlined}, &\quad\text{if } i < T_o \\
       \sum_{i=1}^{T_o - 1} \mathcal{T}_{\Delta \nu}(p)^{(i)}, &\quad\text{if } i = T_o. \\
     \end{cases}
\label{eq:T_delta_nu_p_i}
\end{equation}

\end{document}

%% file: abstract.tex
Haptic guidance is a powerful technique to combine the strengths of humans and autonomous systems for teleoperation. The autonomous system can provide haptic cues to enable the operator to perform precise movements; the operator can interfere with the plan of the autonomous system leveraging his/her superior cognitive capabilities. However, providing haptic cues such that the individual strengths are not impaired is challenging because low forces provide little guidance, whereas strong forces can hinder the operator in realizing his/her plan. 
Based on variational inference, we learn a Gaussian mixture model (GMM) over trajectories to accomplish a given task. The learned GMM is used to construct a potential field which determines the haptic cues. The potential field smoothly changes during teleoperation based on our updated belief over the plans and their respective phases. Furthermore, new plans are learned online when the operator does not follow any of the proposed plans, or after changes in the environment. User studies confirm that our framework helps users perform teleoperation tasks more accurately than without haptic cues and, in some cases, faster. Moreover, we demonstrate the use of our framework to help a subject teleoperate a 7 DoF manipulator in a pick-and-place task.

%% file: intro.tex
Robots are powerful tools for solving tasks that would be very laborious for humans. Furthermore, they can be deployed in hazardous environments, e.g., for sorting nuclear waste~\cite{bloss2010you}. However, especially when operating in unstructured environments, fully autonomous robots may not provide the required reliability. On the other hand, manually controlling the robots, for example via teleoperation, is often cumbersome and inefficient.  
Hence, there is great interest in shared autonomy, where both, the human operator and the autonomous system, influence the robot's actions~\cite{havoutis2019learning}.

\begin{wrapfigure}{R}{0.5\textwidth}
\centering
\includegraphics[width=0.48\textwidth]{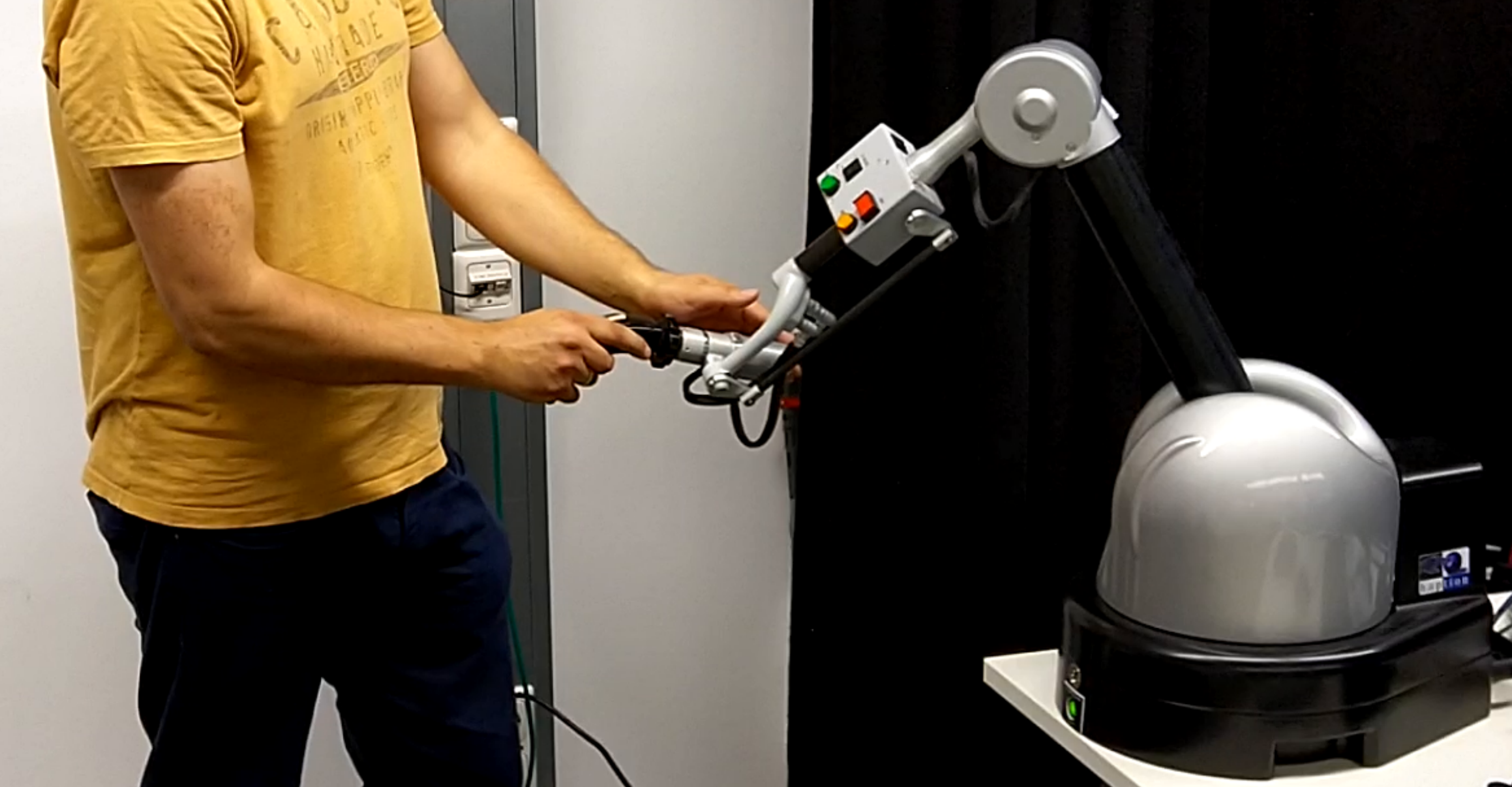}
\caption{User operating the haptic device Haption Virtuose 6D TAO to perform teleoperation tasks.}\label{fig:operator}
\end{wrapfigure}

In safety-critical environments, it is often desirable that the operator remains in full control of the robot. For example, by teleoperation of a robot with a master haptic device~\cite{abi2019haptic}, the autonomous system should apply a wrench on the handle that is large enough to provide guidance, yet weaker than the operator's wrench.
However, the haptic guidance can be inconvenient, especially if the plans of both agents are badly aligned, or if the haptic guidance changes suddenly.

A way of avoiding a misalignment between the plans of the human and of the autonomous system is to vary the stiffness of the master device according to the required precision at each state. This way, the user can easily deviate from the plan proposed by the autonomous system when he/she sees fit as long as this deviation does not compromise the necessary accuracy to solve the task. For example, the robot may need to be controlled with high precision while picking up a small item, whereas the precision while moving towards that item can be rather low since the intention of the human might be still unclear. An autonomous system that takes into account these variabilities while solving a task could apply relatively high wrenches while picking up the small item and restrict itself to merely providing cues while guiding the operator towards that object, enabling the operator to move towards another object for example. In order to encode these variabilities, probabilistic trajectory representations are often used~\cite{ewerton2018assisting, havoutis2019learning}.

\begin{wrapfigure}{R}{0.5\textwidth}
\begin{center}
\includegraphics[width=0.48\textwidth]{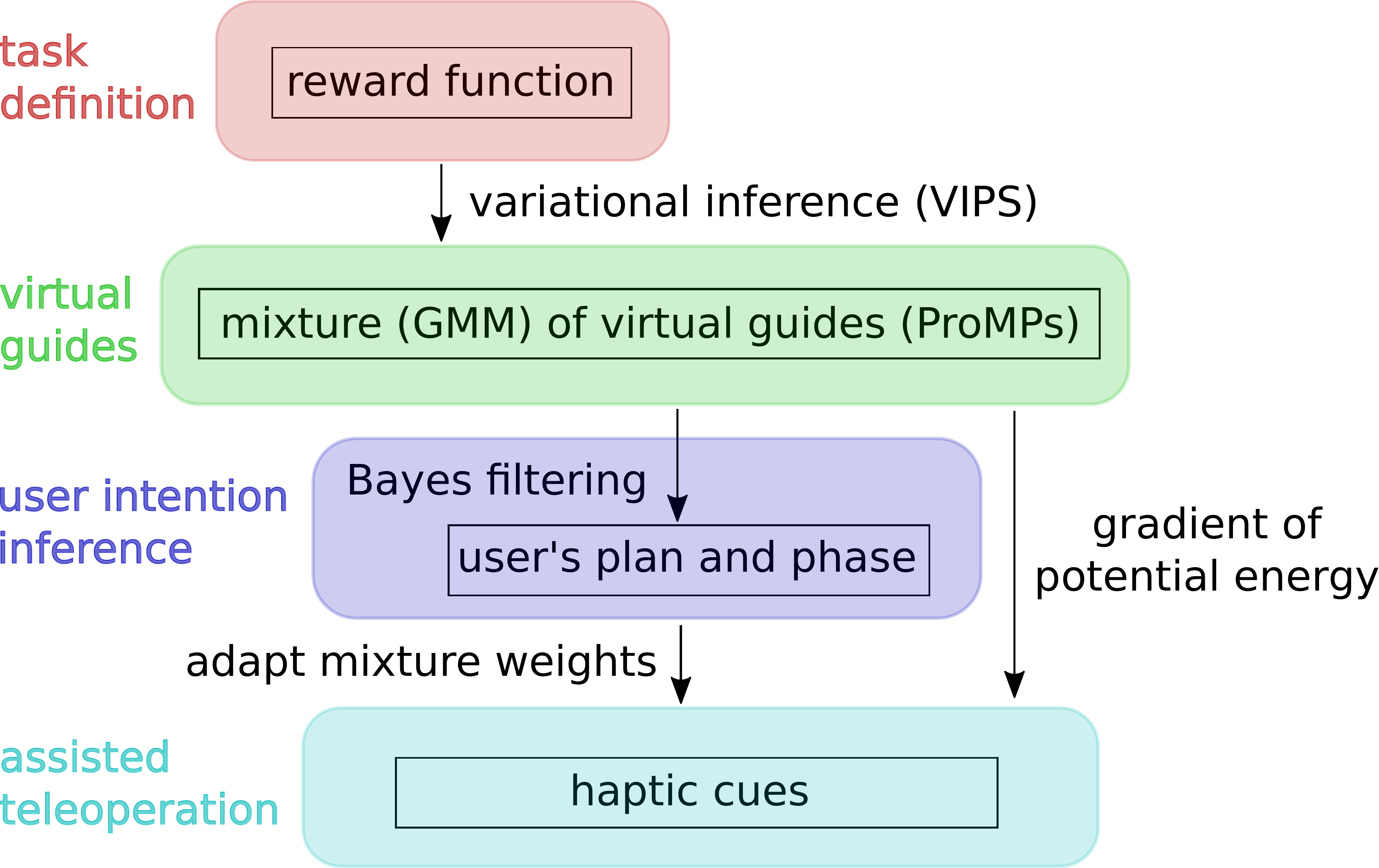}
\end{center}
\caption{Diagram of our proposed framework for assisted teleoperation. First, we define a reward function according to the task at hand (examples in Section~\ref{sec:experiments}). Then, variational inference is used to learn a mixture of virtual guides (Section~\ref{sec:learning_promps_through_VIPS}). Bayes filtering is used to infer the user intention, i.e., the plan and the phase (stage of the movement) intended by the user (Section~\ref{sec:belief_update}). Finally, this inference and the gradient of the potential energy based on the learned mixture modulate the force feedback provided by the haptic device to the user (Section~\ref{sec:haptic_feedback}).}\label{fig:proposed_method_diagram}
\end{wrapfigure}

Gaussian distributions over trajectories, such as probabilistic movement primitives (ProMPs)~\cite{Paraschos2013NIPS}, are particularly convenient because their mean can correspond to a reference trajectory and their covariance matrix can encode the variability along that trajectory. However, a single ProMP only encodes a single (noisy) plan and may thus fail to provide adequate haptic feedback if several plans are feasible. Instead, the autonomous system should consider several possible plans and the possibility of none of its plans being in accordance with the operator's intentions.

We address these challenges by providing haptic feedback based on a mixture of ProMPs, where one of the ProMPs is fixed and has very high variance for all time steps. This ProMP corresponds to a ``freelance'' plan which induces negligible wrenches. The remaining ProMPs are the mixture components of a Gaussian mixture model (GMM) over trajectories which is learned in an episodic maximum entropy reinforcement learning setting. In this work, a method for variational inference, VIPS~\cite{arenz2018efficient}, is used to learn this GMM.

We update our belief over the plan that the operator is following, as well as its phase (the stage of execution of the movement), based on the operator's actions. If the operator does not seem to follow any of the plans of the autonomous system, our framework assigns a high probability to the freelance plan and the operator will receive low haptic feedback. In that case, we can plan a new trajectory online and smoothly blend it in. Furthermore, plans that get invalidated due to changes in the environment can be detected and removed.

In order to provide haptic feedback based on our current belief of the operator's intention, we construct a potential field where each discretized phase of each plan is represented by the energy of a Gaussian distribution in end-effector space and use its gradient to provide haptic feedback. The potential field is thus given by the negative log probability density of a GMM, where the component's weights are given by our belief, and where each plan creates a valley along which the operator is guided. Fig.~\ref{fig:proposed_method_diagram} provides an overview of the modules of our proposed framework and how they are combined to achieve assisted teleoperation.

The efficacy of the proposed method is validated by experiments in which users perform teleoperation tasks by manipulating a haptic device (see Fig.~\ref{fig:operator}). One of the tasks consists of teleoperating a 7 DoF robot arm to pick up objects. The human can switch between different assistive guides and our system helps the user pick up objects accurately. The second task consists of translating and rotating a pole in a virtual environment to reach a certain pose while avoiding obstacles. Both tasks have multiple possible solutions, which are learned by our method.

%% file: connection_to_prior_work.tex
\subsection{Potential Fields and Dynamical Systems}

Potential fields have long found applications in shared control tasks. Aigner and McCarragher~\cite{aigner1997human} use potential fields to drive a robot arm away from obstacles and from the borders of the workspace as well as to let the human change the autonomous behavior of the robot.

Potential fields can be used to ease the teleoperation of a robot to grasp small objects. Howard and Park~\cite{howard2007haptically} describe an experiment where a camera has been mounted on the gripper of a robot. Potential fields help the user move the gripper such that the object appears at the center of the image captured by the camera, meaning that the gripper is just above the object.

Another application of potential fields designed to help users in teleoperation tasks is shown by Gioioso et al.~\cite{gioioso2015force}. In that work, a control loop is proposed to assist the teleoperation of a quadrotor both in contact-free flight and when applying forces to objects.

In contrast to the mentioned works, our approach consists of learning potential fields that provide the user with one or more paths to solve a task. Furthermore, our approach can deal with obstacles without getting stuck in local minima as it is often the case in classical potential fields methods. 

Dynamical systems approaches more sophisticated than classical potential fields have recently been explored in tasks involving avoidance of obstacles of various shapes and interaction with humans. Huber et al.~\cite{huber2019avoidance} have presented an approach to make the end-effector of a robot avoid convex and star-shaped obstacles while moving towards an attractor. That approach consists of multiplying the original linear dynamical system by a dynamic modulation matrix, which can be computed in closed form to guarantee the impenetrability of convex and star-shaped obstacles. Our work does not provide the same strong guarantees. On the other hand, it finds multimodal trajectory distributions to solve planning problems with arbitrary reward functions, which can be used as guiding virtual fixtures in assisted teleoperation tasks.

Potential fields can be learned from demonstrations, which increases the applicability of robots beyond hard-coded behaviors, and can lead to safe and reliable movements in unstructured environments. Khansari-Zadeh and Khatib~\cite{khansari2017learning} propose a method to learn a potential energy function and a dissipative field from demonstrations. The control policy is the negative gradient of the potential function minus the dissipative field. Movements with a single target are addressed. Our work bears similarities to~\cite{khansari2017learning} in the sense that our approach also involves learning a potential field. Nevertheless, in our work, the objective is not necessarily to drive the robot along a certain path to the target. Instead, we address the problem of giving force feedback to a user in assisted teleoperation tasks. There may be several solutions to the task and the user can decide to switch from a path to another. Moreover, our potential field is learned through reinforcement learning instead of from demonstrations. This feature is especially desirable when it is hard for the human to give suitable demonstrations.

\subsection{Gaussian Mixture Models}

Ewerton et al.~\cite{Ewerton2015} propose learning a Gaussian mixture model (GMM) of trajectories from demonstrations to model human-robot collaboration tasks. This approach is extended by Koert et al.~\cite{koert2018online} to enable the open-ended learning of a skill library for collaborative tasks. The open-ended learning is achieved through the use of incremental GMMs~\cite{engel2010incremental}, which do not require pre-specifying the number of mixture components. In this paper, a variational inference method (VIPS) to learn GMMs has been employed~\cite{arenz2018efficient}. VIPS requires specifying a maximum number of components instead of the exact number and is used to learn GMMs given a reward function instead of demonstrations.

Raiola et al.~\cite{raiola2015co} also learn GMMs of trajectories from demonstrations. In their work, virtual mechanisms attract the end-effector of the robot like spring-damper systems, guiding the user in co-manipulation tasks. The GMM formulation allows for the existence of multiple virtual mechanisms. In our work, the GMMs of trajectories are learned through reinforcement learning, which can be very helpful when the user cannot provide suitable demonstrations, e.g. in a challenging teleoperation task with several degrees of freedom.

\subsection{Variational Inference}

Pignat et al.~\cite{pignat2019variational} propose a method to learn a mixture of models through variational inference. In that work, a mixture of Gaussians is learned to cover a certain 2D space while avoiding obstacles. Subsequently, a shortest path algorithm is used to find a sequence of mixture components from a given start position to a given end position. Covering the entire workspace with mixture components does not scale very well for higher dimensions. Therefore, in our work, the components of the learned GMM correspond directly to successful distributions of trajectories to solve the task instead of representing clusters of successful poses.

%% file: learning_ProMPs_through_VIPS.tex
\begin{wrapfigure}{R}{0.5\textwidth}
\centering
\includegraphics[width=0.48\textwidth]{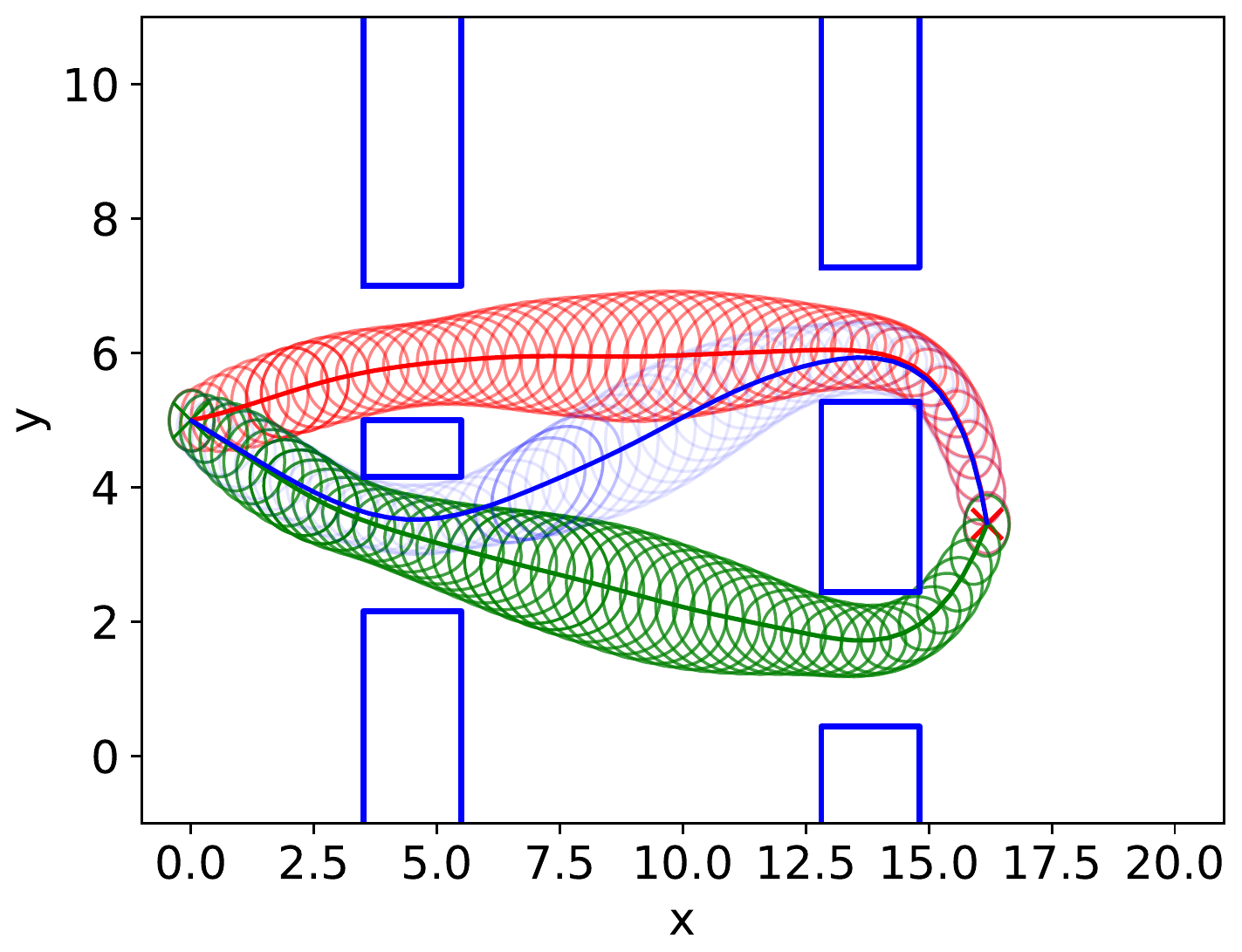}
\caption{Mixture of three trajectory distributions (ProMPs) learned in a 2D scenario. The blue lines represent the sides of walls. A point particle needs to move from a start position (green $\times$) to an end position (red $\times$) while avoiding the walls. VIPS can learn a mixture of ProMPs to solve this task. The ellipses represent Gaussians in Cartesian space, which are used to construct a potential field. This potential field is used to give the user haptic cues as explained in detail in Section~\ref{sec:haptic_feedback}. The number of ellipses, i.e., the number of Gaussians in Cartesian space depends on our phase discretization.}\label{fig:2d_experiment}
\end{wrapfigure}

Teleoperation tasks usually accept multiple solutions. As exemplified in Section~\ref{sec:experiments}, it may be possible to move around obstacles using different plans and there may be multiple targets. Besides, some parts of a successful plan may allow for larger variability than others. In order to encode these variabilities, we want to represent each plan by a trajectory distribution with phase-dependent variance as illustrated in Fig.~\ref{fig:2d_experiment}. We, therefore, represent each plan as a probabilistic movement primitive (ProMP)~\cite{Paraschos2013NIPS}. A ProMP is a Gaussian distribution of trajectories. It can be seen as a Gaussian distribution both in the space of poses as well as in the space of trajectory parameters. This representation is convenient for our needs since the variance of this Gaussian can be used to modulate the stiffness of an autonomous system in a shared autonomy setting, allowing for more or less variability at each stage of a plan. Moreover, ProMPs provide a compact representation of trajectories, which is helpful for reinforcement learning algorithms.

ProMPs represent a trajectory of end-effector poses $\mathbf{x}(\nu) \in \mathbb{R}^n$, where $n$ is the number of degrees of freedom of the end-effector, as a function $\mathbf{x}_{\mathbf{w}}(\nu) = {\boldsymbol{\Phi}(\nu)}\mathbf{w}$ that is linear in normalized Gaussian radial basis functions evenly spaced along the movement phase $\nu \in [0,1]$. The matrix ${\boldsymbol{\Phi}(\nu)} \in \mathbb{R}^{n \times mn}$, where $m$ is the number of basis functions, is a block-diagonal matrix
\begin{equation}
\boldsymbol{\Phi}(\nu)=  \begin{bmatrix}
    \boldsymbol{\phi}(\nu)^{\top} & & \\
    & \ddots & \\
    & & \boldsymbol{\phi}(\nu)^{\top}\setlength{\multlinegap}{0pt}
  \end{bmatrix}.
\label{eq:Phi_nu}
\end{equation}
Each element of the vector $\boldsymbol{\phi}(\nu) \in \mathbb{R}^m$ is a normalized Gaussian radial basis function of the form
\begin{equation}
\phi_c(\nu) = \dfrac{\exp \left( - \frac{1}{2} \left( \nu - \frac{c - 1}{m-1} \right)^2 \right)}{\sum\limits_{c=1}^{m} \exp \left( - \frac{1}{2} \left( \nu - \frac{c - 1}{m-1} \right)^2 \right)}
\label{eq:g}
\end{equation}
evaluated at $\nu$, where the variable $c \in \{1, 2, \cdots, m\}$ specifies the center of each basis function.

The trajectory distributions (ProMPs) depicted in Fig.~\ref{fig:2d_experiment}, for instance, have $m=10$ radial basis functions and $n=2$ in this case. Since the pose $\mathbf{x}_{\mathbf{w}}(\nu)$ is an affine function of the weight vector $\mathbf{w} \in \mathbb{R}^{mn}$, a Gaussian distribution over weights $p(\mathbf{w})=\mathcal{N}(\mathbf{w}|\boldsymbol{\mu}_{w}, \boldsymbol{\Sigma}_{w})$ induces a Gaussian distribution over poses 
\begin{align}
    p(\mathbf{x}|\nu) &= \mathcal{N}\left(\mathbf{x}|\boldsymbol{\Phi}(\nu) \boldsymbol{\mu}_{w},  {\boldsymbol{\Phi}(\nu)} \boldsymbol{\Sigma}_{w} \boldsymbol{\Phi}(\nu)^\top\right).
\label{eq:p_x_given_nu}
\end{align}
Hence, a mixture of ProMPs can be represented by a GMM over weights
\begin{equation}
    p(\mathbf{w}) = \sum_{o=1}^{N_o} p(o) p(\mathbf{w}|o),
\label{eq:p_w}
\end{equation}
where a multinomial distribution $p(o)$ assigns a normalized weight to each option $p(\mathbf{w}|o)=\mathcal{N}(\mathbf{w}|\boldsymbol{\mu}_o, \boldsymbol{\Sigma}_o)$ and $N_o$ is the number of options. Each option represents a different plan. Figs.~\ref{fig:2d_experiment}~and~\ref{fig:real_robot_3D_experiments}, for example, show situations where three options have been learned.

We consider a reinforcement learning setting with a reward function $r(\mathbf{x}_{\mathbf{w}}(\nu),\nu)$ that depends on the end-effector pose $\mathbf{x}$ and the phase $\nu$. Such a reward function can, for example, assume higher values when close to desired poses or when the acceleration is low and lower values when close to obstacles or when the acceleration is high. We aim to learn a mixture of ProMPs that achieves high episodic reward $r(\mathbf{w}) = \int_{0}^{1} r(\mathbf{x}_{\mathbf{w}}(\nu),\nu) d\nu$ in expectation while maintaining high entropy. These assumptions can be phrased as the optimization problem
\begin{equation}
\label{eq:optimizationProblem}
\underset{p(o), \boldsymbol{\mu}_{o}, \boldsymbol{\Sigma}_o}{\arg\max} \sum_{o=1}^{N_o} p(o) \int p(\mathbf{w}|o) r(\mathbf{w}) d\mathbf{w} + H(p(\mathbf{w}))    
\end{equation}
with the Shannon entropy $H(p(\mathbf{w}))$ of the GMM. The entropy objective rewards variability and is crucial to prevent convergence to a single trajectory. Please note that scaling the reward function affects the optimal solution due to the entropy objective. We assume that the reward function is adequately scaled to result in sufficient variability without inducing undesirable trajectories.

As we consider an episodic setting, that is, we do not assume nor exploit access to time-series data, the maximum entropy reinforcement learning problem (\ref{eq:optimizationProblem}) can be equivalently framed as variational inference.
The connection between this reinforcement learning formulation and variational inference can be shown by introducing a reward distribution
\begin{equation}
p_r(\mathbf{w}) = \frac{1}{Z_r} \exp(r(\mathbf{w}))
\label{p_r_w}
\end{equation}
with partition function $Z_r = \int \exp(r(\mathbf{w})) d\mathbf{w}$.
The optimization problem (\ref{eq:optimizationProblem}) can now be reformulated as the variational inference problem of approximating the reward distribution $p_r(\mathbf{w})$ with a GMM $p(\mathbf{w})$ by minimizing the reverse Kullback-Leibler divergence $D_\text{KL}(p(\mathbf{w})||p_r(\mathbf{w}))$. This optimization problem is
\begin{align}
    \label{eq:vi_optimizationProblem}
    &\underset{p(o), \boldsymbol{\mu}_{o}, \boldsymbol{\Sigma}_o}{\arg\min} D_\text{KL}(p(\mathbf{w})||p_r(\mathbf{w})) \nonumber \\
    &=\underset{p(o), \boldsymbol{\mu}_{o}, \boldsymbol{\Sigma}_o}{\arg\max} \int p(\mathbf{w}) \log{p_r(\mathbf{w})} d\mathbf{w} + H(p(\mathbf{w})) \nonumber \\
    &=\underset{p(o), \boldsymbol{\mu}_{o}, \boldsymbol{\Sigma}_o}{\arg\max} \int p(\mathbf{w})r(\mathbf{w}) d\mathbf{w} - \log Z_r + H(p(\mathbf{w})).
\end{align}
As the constant partition function $Z_r$ does not affect the solution of the optimization problem, (\ref{eq:optimizationProblem}) and (\ref{eq:vi_optimizationProblem}) are equivalent. Therefore, we can learn a mixture of ProMPs with any variational inference method that is capable of learning GMMs. We use variational inference by policy search (VIPS)~\cite{arenz2018efficient}. Fig.~\ref{fig:2d_experiment} shows a mixture of three ProMPs learned by VIPS to solve a planning problem in a 2D scenario.

%

%% file: haptic_feedback.tex
Haptic cues to the operator are provided by a potential field in end-effector space based on our GMM in weight space. 
More precisely, we discretize the phase $\nu$ in $T_o$ steps for each plan, denoted by $\nu_1, \dots \nu_{T_o}$, and compute the marginal distribution
\begin{align}
    &p(\mathbf{x}) = \sum_{o=1}^{N_o} p(o) \sum_{i=1}^{T_o} p(\nu_i|o) p(\mathbf{x}|\nu_i, o) \nonumber \\
    &= \sum_{o=1}^{N_o} p(o) \sum_{i=1}^{T_o} p(\nu_i|o) \mathcal{N}(\mathbf{x}|\boldsymbol{\Phi}(\boldsymbol{\nu}_i) \boldsymbol{\mu}_o, {\boldsymbol{\Phi}(\nu_i)} \boldsymbol{\Sigma}_{o} \boldsymbol{\Phi}(\nu_i)^\top) \nonumber \\
    &= \sum_{o=1}^{N_o} p(o) \sum_{i=1}^{T_o} p(\nu_i|o) \mathcal{N}(\mathbf{x}|\boldsymbol{\mu}_{o, i}, \boldsymbol{\Sigma}_{o, i})
\label{eq:p_x}
\end{align}
using our belief $p(\nu_i|o)$ about the current phase of each plan $o$ and the distribution $p(\mathbf{x}|\nu_i, o)$ over end-effector poses at phase $\nu_i$ in plan $o$. Each ellipse in Fig.~\ref{fig:2d_experiment} corresponds to a distribution $p(\mathbf{x}|\nu_i, o)$. We will discuss in Section \ref{sec:belief_update} how we adapt online the belief $p(\nu_i|o)$ over the phase and also the belief $p(o)$ over the plan.

The marginal distribution $p(\mathbf{x})$ corresponds to a GMM with components for each plan and each phase. Similarly to Hamiltonian MCMC~\cite{bishop2006pattern}, we define a potential energy $E(\mathbf{x}) = - \log p(\mathbf{x})$ based on the negative log of the marginal, which induces a wrench that is given by the negative gradient of the energy
\begin{align}
    \boldsymbol{\tau}(\mathbf{x}) &= - \frac{\partial E(\mathbf{x})}{\partial \mathbf{x}} = \frac{\partial \log p(\mathbf{x})}{\partial \mathbf{x}}.
\label{eq:tau_x}
\end{align}
By applying the wrench $\boldsymbol{\tau}(\mathbf{x})$, we guide the operator towards low energy regions based on the marginal $p(\mathbf{x})$. As the marginal is computed based on an approximation of the episodic reward distribution $p_r(\mathbf{w})$, the operator is guided towards trajectories that provide a high episodic reward. Please note that directly using the negative reward function $-r(\mathbf{x}, \nu)$ for constructing a potential field would guide the operator towards regions of high immediate reward, whereas our learned, planning-based potential field assists the operator in achieving a high episodic reward. For example, consider a goal position that is directly behind a wall and a reward function that penalizes the distance to the goal and close proximity to the wall. Whereas the immediate reward results in a potential field that would keep the operator in the current position, our learned potential field would create several valleys along which the operator is guided around the wall. 
Furthermore, the energy of the marginal distribution $p(\mathbf{x})$ corresponds to the negative log of a GMM and is thus differentiable and smooth. A smooth potential field is highly desirable in order to avoid sudden jumps in the direction or magnitude of the wrench that is applied to the handle of the master device.
The gradient of the log of the marginal is
\begin{equation}
    \label{eq:gradientOfGMM}
    \frac{\partial \log p(\mathbf{x})}{\partial \mathbf{x}} = \sum_{o=1}^{N_o} \sum_{i=1}^{T_o} p(o, \nu_i | \mathbf{x}) \Sigma_{o, i}^{-1} (\mu_{o, i} - \mathbf{x}).
\end{equation}
Hence, each component attracts the operator with a force of magnitude proportional to the distance to its mean and weighted by its responsibility
\begin{equation}
p(o, \nu_i | \mathbf{x}) = \frac{p(o) p(\nu_i|o) p(\mathbf{x}|\nu_i, o)}{p(\mathbf{x})}.
\label{eq:p_o_nu_i_given_x}
\end{equation}
The operator would typically get small haptic cues if he/she is close to one of the components because, due to the proximity to that component, its responsibility would often be close to one and its force contribution would be close to zero. However, if the operator is far from every component, all components would have large force contributions and their weighted average would typically also be large. Such behavior can be desirable if we do not want to allow the operator to leave the planned trajectories. 

However, if the operator should keep full control, such behavior would be undesirable even if the maximum wrench gets capped. In such cases, we can add an additional component with large variance to our mixture model that covers the whole workspace. Such a component corresponds to a freelance plan with a single phase, where the operator can move freely within the workspace. Due to its high variance, this component always has low force contributions and gets low responsibility if the operator is close to the planning based components, but high responsibility if the operator is far from any planned components. 

Without the user, when assuming the beliefs $p(o)$ and $p(\nu_i|o)$ to be fixed and the control rate at which the haptic cues are updated to be infinite, no energy gets injected into our system. We include a damping term, such that the total wrench is
\begin{equation}
    \boldsymbol{\tau}_\text{total}(\mathbf{x}, \mathbf{\dot{x}}) = \boldsymbol{\tau}(\mathbf{x}) - k_\text{damp}  \mathbf{\dot{x}}.
\end{equation}
By including such a damping, the total energy of the system always decreases, which leads to its stability. The damping term also prevents oscillatory force feedback close to low-variance components that were caused by our limited control rate.

%% file: belief_update.tex
In order to assist the user in a teleoperation task, our system needs to infer from the pose of the handle of the haptic device commanded by the human which ProMP (also referred to as option or plan) the human intends to follow. We may also want to guide the human forward along the intended plan while enabling him/her to stop or move backward. Therefore, we may also need to infer the intended phase along the plan of interest.

These inference problems are formalized by using the weights $p(o)$ and $p(\nu|o)$ to reflect our belief about the plan $o$ that the operator is currently employing and about the phase $\nu$ along each plan. We initialize $p(o)$ based on the weights that were learned by VIPS, which assigns higher weights to components that achieve a higher expected reward, higher entropy or that differ from the remaining components. We initialize our belief $p(\nu|o)$ over the phase as a uniform distribution.

\begin{figure}
\centering
\includegraphics[width=\textwidth]{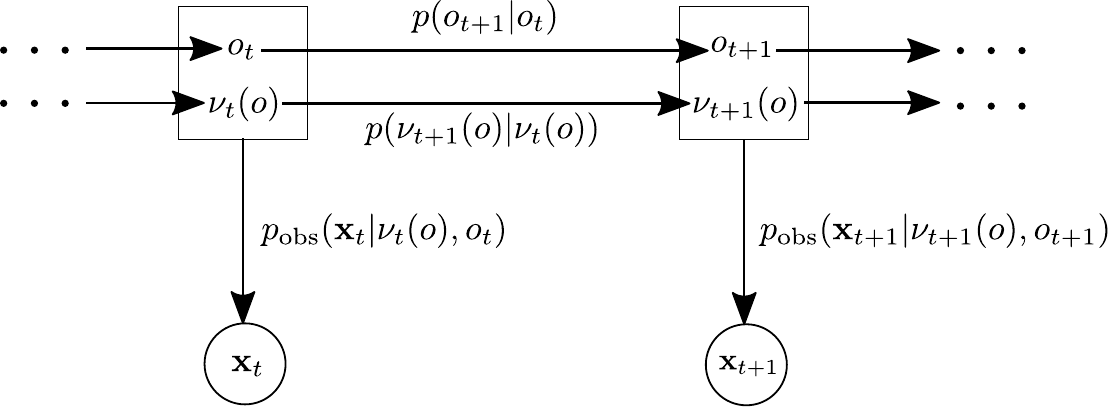}
\caption{Hidden Markov model (HMM) underlying the update of the belief about the plan $o$ and phase $\nu$.}
\label{fig:hmm}
\end{figure}

In order to update our belief during teleoperation, we formulate a hidden Markov model (HMM) where the true plan $o_t$ of the operator and the phases ${\nu}_t(o)$ at time $t$ are hidden variables. Please note that $t$ corresponds to the discretized real time that increments at the control rate at which we send the desired wrench to the haptic device. Fig.~\ref{fig:hmm} represents this HMM. The transition probabilities $p(\nu_{t+1}(o)|\nu_t(o))$ and $p(o_{t+1}|o_{t})$ are explained in Sections~\ref{sec:updating_the_belief_about_the_phase}~and~\ref{sec:updating_the_belief_about_the_plan}, respectively. The emission probability $p_{\text{obs}}(\mathbf{x}_{t}|\nu_{t}(o), o_{t})$ is obtained by rescaling the Covariance matrix of $p(\mathbf{x}_{t}|\nu_{t}(o), o_{t})$ to allow for a larger uncertainty in the observation $\mathbf{x}_{t}$. The rescaling factor for the Covariance matrix is chosen such that the belief about the plan $p(o)$ does not converge too fast in the face of new observations.

\subsection{Updating the Belief about the Phase}
\label{sec:updating_the_belief_about_the_phase}
\begin{figure}
\centering
\includegraphics[width=\textwidth]{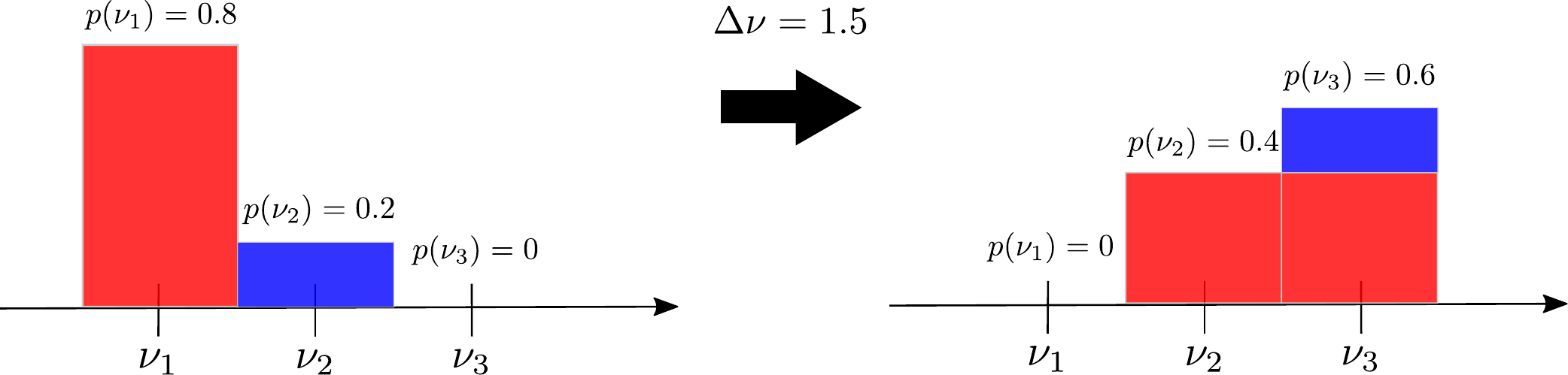}
\caption{Example of change in the probability distribution $p(\nu)$ induced by the shifting operator $\mathcal{T}_{\Delta\nu}(p)$ when the phase is discretized in only three possible values for simplification. With $\Delta \nu = 1.5$, half of the probability mass of $\nu_1$ gets assigned to $\nu_2$ and the other half gets assigned to $\nu_3$. Because $\nu_3$ is the last phase, all the probability mass of $\nu_2$ gets shifted to $\nu_3$.}
\label{fig:transition_by_delta_nu}
\end{figure}

For modeling the transition probability of the phase $p(\nu_{t+1}(o)|\nu_t(o))$, we assume that the phase progresses by $\Delta\nu$ discrete steps with probability $p_\text{progress}$ and changes to a random phase otherwise. A prior belief about the phase at time step $t$ can thus be computed based on the posterior belief at time step $t-1$,
\begin{align}
    \label{eq:weightTransition}
    p_\text{prior,t}(\nu|o) = &p_\text{progress} \mathcal{T}_{\Delta\nu}\big(p_\text{post,t-1}(\nu|o)\big) + (1-p_\text{progress}) \frac{1}{T_o},
\end{align}
where the shifting operator $\mathcal{T}_{\Delta\nu}(p)$ shifts the probability mass of the distribution $p$ by $\Delta\nu$ forwards in time whereby no mass gets shifted beyond the last phase $\nu_{To}$. Please note that $\Delta\nu$ can be any positive real value by shifting fractions of the probability mass of the distribution $p$. For example, when shifting by $\Delta\nu=1.5$, half of the probability at $\nu_1$ would get assigned to $\nu_2$ and the other half would get assigned to $\nu_3$. When shifting by $\Delta\nu=2.0$, the whole probability at $\nu_1$ would get assigned to $\nu_3$. Fig.~\ref{fig:transition_by_delta_nu} presents a visualization of the effects of the shifting operator $\mathcal{T}_{\Delta\nu}(p)$ for a case where $T_o = 3$. A formal definition of $\mathcal{T}_{\Delta\nu}(p)$ is provided in Appendix~\ref{sec:definition_of_the_shifting_operator}. 

Please note that the progress of the continuous phase, $\delta_\nu(T_o) = \frac{\Delta_\nu}{T_o}$, also depends on the number of discretizations for a given plan. Hence, the progress for a given plan can be individually controlled by changing the number of discretizations, $T_o$. For all our experiments, we chose the same number of discretizations for all plans (except for the stationary freelance-plan) because we do not want different plans to progress with different speed. However, it would be a natural extension to adapt the number of discretizations, for example, depending on the length of the planned (mean) trajectory.

Based on recursive Bayesian estimation, the posterior belief at time $t$ is
\begin{align}
p_\text{post,t}(\nu|o) &\propto p_\text{prior,t}(\nu|o) p_\text{obs}(\mathbf{x}_t|\nu, o).
\label{eq:p_post_t_nu_given_o}
\end{align}
However, when providing haptic cues based on the posterior belief of the phase at time step $t$, we would not provide any incentive to the operator to progress with the plan. Instead, we compute the haptic cues based on the prior belief of the phase at the next time step, $p_\text{prior,t+1}(\nu|o)$. 
As we assume the phase to progress by $\Delta\nu$ discrete steps at the control rate, computing the negative energy gradient (\ref{eq:gradientOfGMM}) based on the prior belief of the next time step results into a wrench that tends to pull the operator towards the upcoming components for each plan, providing an incentive to progress.

\subsection{Updating the Belief about the Plan}
\label{sec:updating_the_belief_about_the_plan}
We model the transition probability of the plan as  
\begin{align}
    p(o|o') =  \left\{
                \begin{array}{ll}
                  (1-p_\text{switch}) \text{, if }o'= o \\
                  p_\text{switch}/(N_o - 1) \text{, otherwise}\\                \end{array}
              \right.
\end{align}
with a small probability $p_\text{switch}$ of changing the plan and $N_o$ corresponding to the number of plans. We therefore assume that the operator likely follows the same plan as during the last time step, but with low probability might change to another random plan.
The Bayesian belief update
\begin{align}
    p_\text{post,t}(o) & \propto p_\text{prior,t}(o) p(\mathbf{x}_t|o) \nonumber \\ 
    & \propto \int p_{t-1}(o')  p(o|o') do' \sum_{i=1}^{T_o} p_t(\nu_i|o) p_{\text{obs}}(\mathbf{x}_t|\nu_i, o)
\end{align}
corresponds to the posterior belief of the plan that is pursued by the operator at time $t$.

%% file: online_adaption.tex
The computational time required to plan a mixture of ProMPs with the proposed method depends strongly on the number of degrees of freedom and basis functions, as well as on the number of components $p(\mathbf{x}|o)$ that are to be learned. VIPS~\cite{arenz2018efficient} can learn components with full covariance matrices $\Sigma_o$; however, to speed up the online replanning of ProMPs, we only learn diagonal covariance matrices. This simplification ignores correlations between different phases and between different degrees of freedoms, resulting in mixture components $p(\mathbf{x}_t|\nu, o)$ in the end-effector space that are axis-aligned.

We found VIPS to be sufficiently efficient for learning small mixtures of ProMPs online. We consider two kinds of events that trigger online replanning, namely, replanning due to changes in the environment and replanning if the operator does not seem to pursue any of the existing plans. 

Changes of the environment can be detected for example via vision and typically invalidate the previous plans. We, therefore, remove the previous plans and provide haptic cues only based on the freelance plan and the new plans.

We can detect that the operator is not pursuing any of the current plans based on the belief $p(o_\text{freelance})$ of the freelance plan. If the operator does not seem to pursue any of the previous plans, we do not need to delete the previous plans. In order to avoid sudden changes in the haptic cues, we add the new plans with negligible initial weight to the mixture of ProMPs. The weight of the new plans will typically smoothly increase due to the belief updates.

%% file: experiments.tex
Two experiments demonstrate applications of our framework to assist humans to perform teleoperation tasks. The haptic device used in these experiments, the Virtuose 6D TAO manufactured by the company Haption \cite{Virtuose6D_TAO}, is depicted in Fig.~\ref{fig:operator}. The first experiment shows that our framework can help users teleoperate a 7 DoF robot arm. In the second experiment, users need to control the 6D pose of a pole in a virtual environment. User studies validate the efficacy of our framework. The user studies have been conducted in the second experiment to avoid the risk of damaging real equipment as it could be the case due to collisions with objects in the environment of the teleoperated robot arm. Future user studies shall be conducted in tasks involving the teleoperation of real robot arms.

The ProMPs used in these experiments have $m=7$ basis functions. We have chosen a number of basis functions low enough to enable a reinforcement learning algorithm to quickly find successful ProMPs and high enough to express the diversity of trajectories that satisfy the criteria of our tasks, such as avoiding obstacles and reaching desired targets. In the experiment described in Section~\ref{sec:3d_experiment}, the number of degrees of freedom controlled by the user is $n=3$, while, in Section~\ref{sec:6d_experiment}, $n=6$. See Section~\ref{sec:learning_promps_through_VIPS} for the definition of ProMPs.

In the experiment described in Section~\ref{sec:3d_experiment}, $p_\text{progress}=0.8$ and $\Delta\nu=0$, which means that, in that experiment, although the user was constrained by the learned guiding ProMPs, there was no force pulling him/her towards any of the targets. This made it convenient for the user to pause the movement while controlling the real robot. In Section~\ref{sec:6d_experiment}, $p_\text{progress}=0.8$ and $\Delta\nu=0.5$, i.e., there was a small force pulling the user towards the target. See Section~\ref{sec:updating_the_belief_about_the_phase} for the explanation of the parameters $p_\text{progress}$ and $\Delta\nu$. In both experiments described in this section, $p_\text{switch} = 1\mathrm{e}{-20}$ (see Section~\ref{sec:updating_the_belief_about_the_plan}).

\begin{figure}
\centering
\subfloat[Initial virtual guides (left) and environment (right)]{%
\resizebox*{\textwidth}{!}{\includegraphics{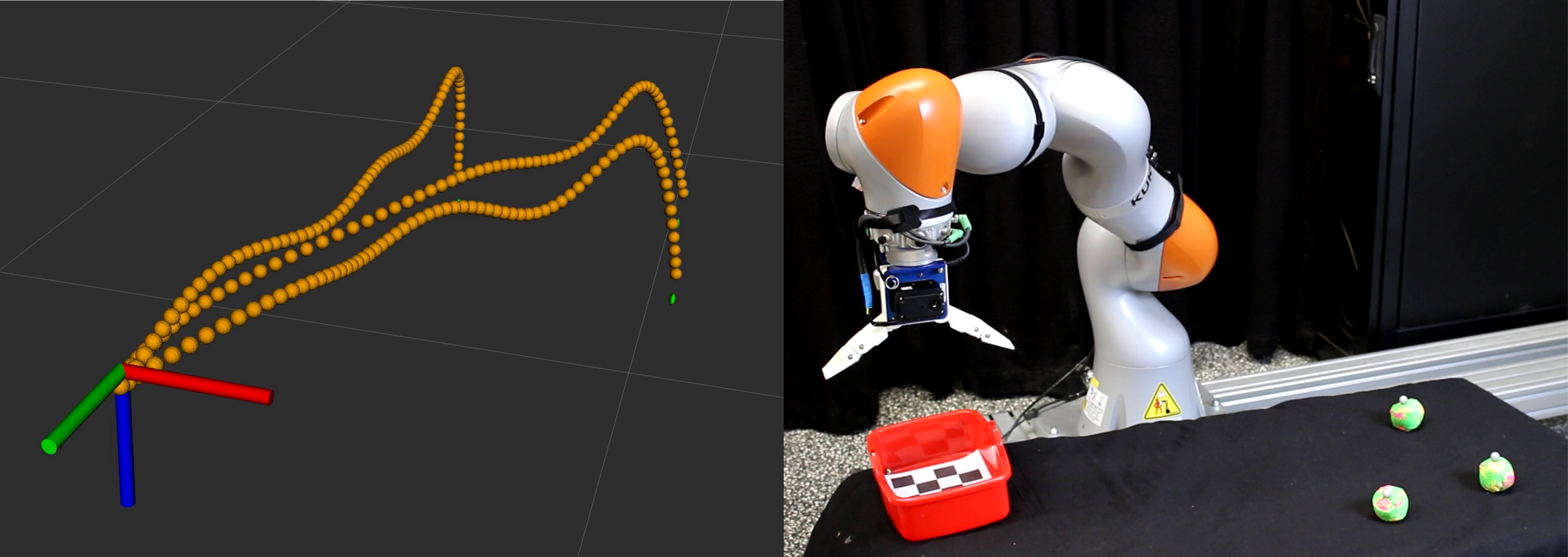}}}\vspace{5pt}
\subfloat[Virtual guides (left) and environment (right) after first grasp]{%
\resizebox*{\textwidth}{!}{\includegraphics{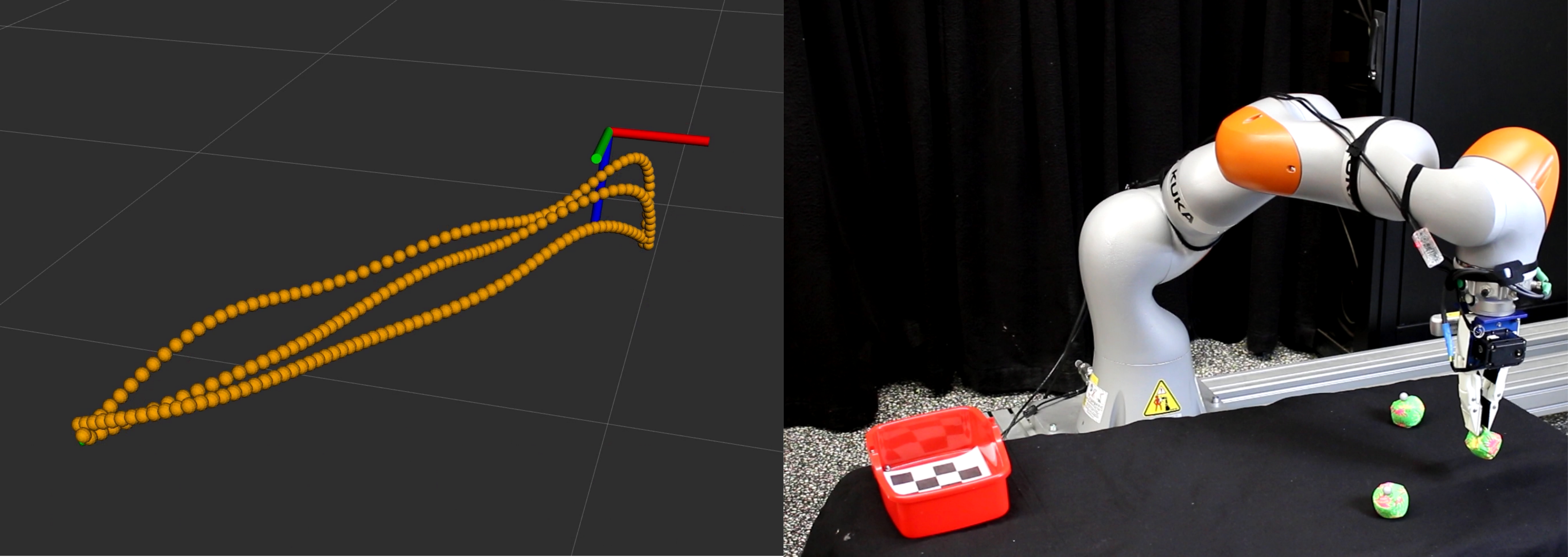}}}\vspace{5pt}
\subfloat[Virtual guides (left) and environment (right) after introducing an obstacle]{%
\resizebox*{\textwidth}{!}{\includegraphics{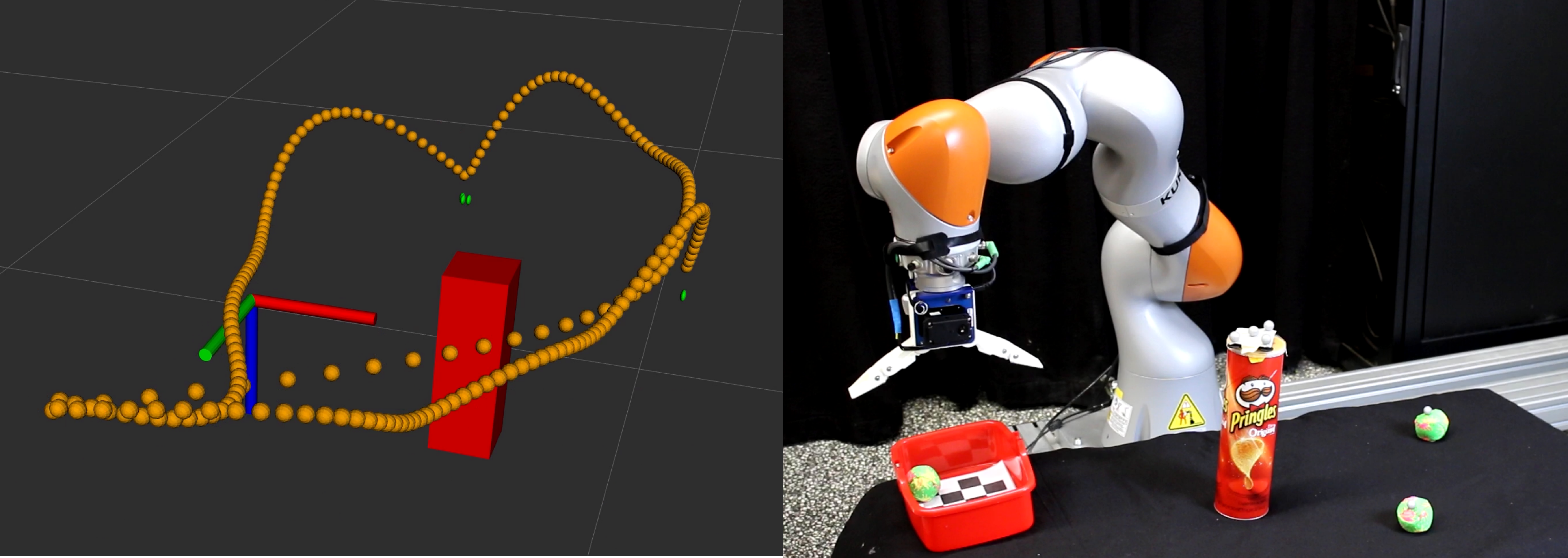}}}
\caption{Teleoperating a robot arm through a haptic device with force feedback. In this experiment, the user controls the Cartesian position of the robot end-effector. The objective of the user is to put all the green balls inside the red basket while avoiding the obstacle (a Pringles can). Our framework computes distributions of trajectories that serve as virtual guides to the user. The force feedback applied by the haptic device on the user depends on the gradient of the trajectory distributions. Our framework replans the virtual guides whenever there is a change in the environment or when the user escapes all previously planned virtual guides.}
    \label{fig:real_robot_3D_experiments}
\end{figure}

\subsection{Shared Control of a Robot Arm}
\label{sec:3d_experiment}

Picking up small objects through teleoperation can be a very challenging task for a human operator due to the difficulty in accurately estimating the 3D position of the objects of interest. The purpose of this experiment is to verify if our framework can learn multiple virtual guides to help a user pick up small objects by teleoperating a robot arm while avoiding obstacles. This experiment also tests if helpful virtual guides can be learned online in the face of changes in the environment, such as changes in the positions of the target objects and of the obstacle.

In our setup, the human manipulates a 6 DoF haptic device \cite{Virtuose6D_TAO} to teleoperate a 7 DoF robot arm, the LBR iiwa 14 R820 manufactured by KUKA \cite{iiwa}. The human has control over the 3D Cartesian end-effector position of the robot arm and the joint configuration of the robot is computed through inverse kinematics. The goal of the human is to pick three little balls and release them inside a basket. The environment may change online through the introduction of an obstacle (a Pringles can). Our framework helps the user solve this task by learning multiple plans (ProMPs) that modulate the force feedback applied by the haptic device on the human. The ProMPs are relearned online whenever there are changes in the environment or when the belief of the freelance plan is larger than $0.5$. Fig.~\ref{fig:real_robot_3D_experiments} shows three snapshots during one trial to accomplish the pick and place task. The snapshots show that the virtual guides are learned online when the environment changes and that they are in accordance with the objectives of the task, which are reaching each of the balls to put them in the red basket and avoiding the obstacle (a Pringles can). The balls and the obstacle have been tracked by using the OptiTrack Flex 13 system~\cite{OptiTrack}.

In these experiments, the reward $r(\mathbf{w}) = \boldsymbol{\psi}(\mathbf{x_w})^{\top} \boldsymbol{\theta}$ is a weighted sum of features. The vector $\boldsymbol{\theta}$ has empirically determined weights for each feature in $\boldsymbol{\psi}(\mathbf{x_w})$. We have manually determined these weights by verifying if the distributions learned in a virtual model of our environment generated successful trajectories. An alternative to empirically determining the weights would be using inverse reinforcement learning (IRL)~\cite{abbeel2004apprenticeship}, which shall be investigated in our future research.

In this experiment, $\mathbf{x_w}(\nu_i) = [x(\nu_i), y(\nu_i), z(\nu_i)]^\top$ represents the 3D Cartesian position of the end-effector of the slave robot arm at phase $\nu_i$. Our feature vector evaluated for trajectory $\mathbf{x}$ is
\begin{equation}
    \boldsymbol{\psi}(\mathbf{x}) = \left[ d_{\text{start}}^2, d_{{\text{end}}}^2, \mathcal{L}_{\text{box}}, \mathcal{L}_{\text{obstacle}}, \sum_{i=1}^{T_o}\mathbf{\dot{x}}^2(\nu_i), \sum_{i=1}^{T_o}\mathbf{\ddot{x}}^2(\nu_i), \sum_{i=1}^{T_o} \min (0.5, z(\nu_i)) \right]^\top
    \label{eq:reward_features_3D_experiment}
\end{equation}
and the vector of feature weights is $\boldsymbol{\theta} = [-5000, -5000, 5000, 5000, -500, -50000, 50]^\top$.

In~(\ref{eq:reward_features_3D_experiment}), $d_{\text{start}}$ is the distance between the start position of the trajectory and the desired start position. The variable $d_{\text{end}}$ represents the distance between the end position and its closest target. The features $\mathcal{L}_{\text{box}}$ and $\mathcal{L}_{\text{obstacle}}$ prevent leaving the workspace in the form of a bounding box and collisions with the obstacle, respectively. The next two features prevent high velocities and accelerations. Finally, the last feature gives higher rewards for achieving high positions inside the bounding box defining the workspace of the robot. This feature induces grasping motions from the top.

The features $\mathcal{L}_{\text{box}}$ and $\mathcal{L}_{\text{obstacle}}$ are computed based on the signed Euclidean distance between the robot end-effector and the bounding box or obstacle, respectively. First, the minimum signed Euclidean distance $d$ between trajectory $\mathbf{x}$ and the bounding box or obstacle is computed. Subsequently, the log-likelihood of that trajectory is given by
\begin{equation}
	 \mathcal{L} = \left\{ 
	 			\begin{array}{ll}
                  \log\left(\mathcal{N}\left(0; 0, \sigma^2\right)\right) \text{, if } d \geq 0  \\
                   \log\left(\mathcal{N}\left(d; 0, \sigma^2\right)\right) \text{, if } d < 0
                \end{array}
              \right.
    \label{eq:L_obstacle}
\end{equation}
where we chose $\sigma^2 = 2$. This choice was based on our experience in simulation. By $\mathcal{N}\left(0; 0, \sigma^2\right)$ and $\mathcal{N}\left(d; 0, \sigma^2\right)$, we denote the probability density function of a univariate Gaussian with mean 0 and standard deviation $\sigma$ evaluated at locations 0 and $d$, respectively. The signed Euclidean distance to the bounding box is positive inside the box and negative outside it. The obstacle is modeled as a cylinder. The signed Euclidean distance to the obstacle is positive outside the cylinder and negative inside it.

\subsection{Shared Control of a Virtual Object}
\label{sec:6d_experiment}

\begin{wrapfigure}{R}{0.5\textwidth}
\begin{center}
\includegraphics[width=0.48\textwidth]{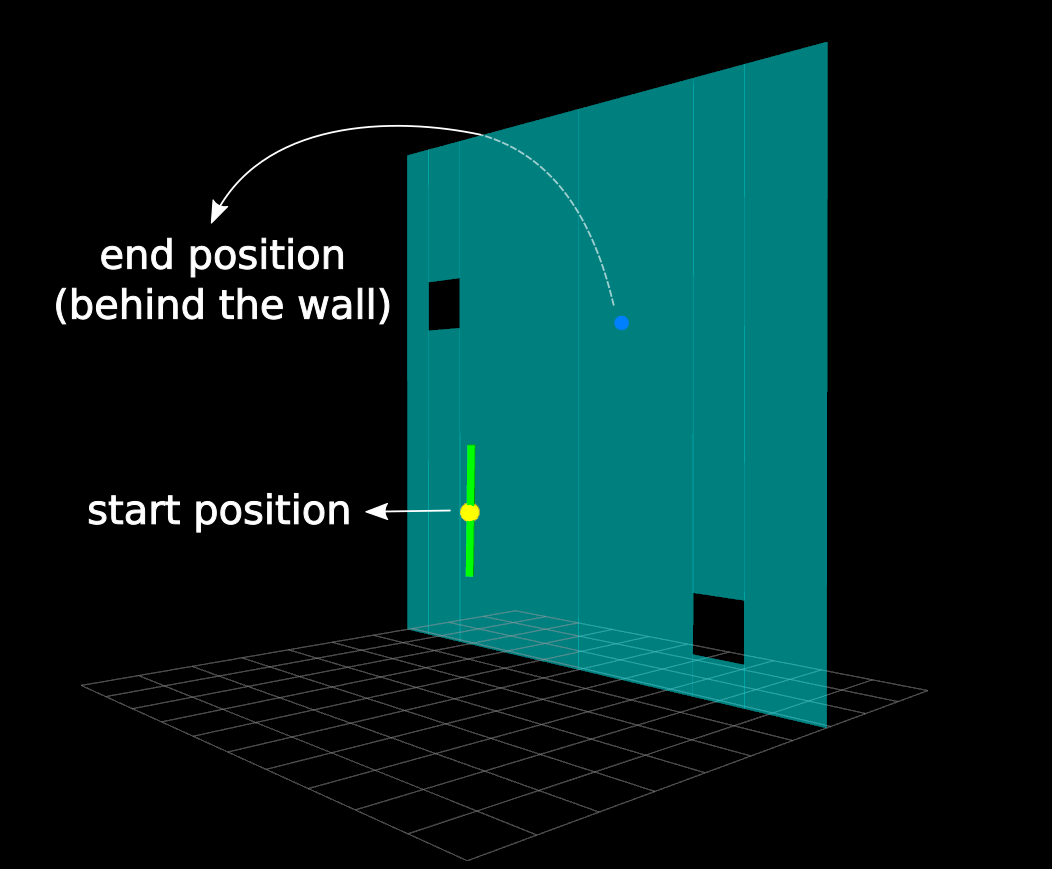}
\end{center}
\caption{6D Experiment. The user operates the haptic device Haption Virtuose 6D TAO to translate and rotate a pole in a virtual environment. The goal of the user is to move the green pole from the start position (yellow sphere) to the end position (blue sphere) through one of the two windows without hitting the wall. VIPS learns a mixture of trajectory distributions (ProMPs) that helps the user achieve the necessary translations and rotations to solve this task.}\label{fig:6d_experiment}
\end{wrapfigure}

This experiment addresses assisting teleoperation with multiple solutions involving both translations and rotations. The main purpose of this experiment is to statistically test whether or not, in comparison with not using force feedback, our framework helps users avoid collisions and complete teleoperation tasks faster while retaining control over the haptic device. Here, users control a virtual pole, which allows for a simpler setup to perform user studies and minimizes the risk of damaging real equipment. Fig.~\ref{fig:6d_experiment} depicts our virtual environment. As in the previous experiment, the reward function has the form $r(\mathbf{w}) = \boldsymbol{\psi}(\mathbf{x_w})^{\top} \boldsymbol{\theta}$. This time, the feature vector for trajectory $\mathbf{x}$ is
\begin{equation}
    \boldsymbol{\psi}(\mathbf{x}) = \left[ d_{\text{start}}^2, d_{{\text{end}}}^2, \mathcal{L}_{\text{wall}}, \sum_{i=1}^{T_o}\mathbf{\dot{x}}^2(\nu_i), \sum_{i=1}^{T_o}\mathbf{\ddot{x}}^2(\nu_i), \sum_{i=1}^{T_o} \alpha^2(\nu_i) \!+\! \beta^2(\nu_i) \!+\! \gamma^2(\nu_i) \right]^\top
    \label{eq:reward_features_6D_experiment}
\end{equation}
and the vector of feature weights is $\boldsymbol{\theta} = [-2.5, -5, 1000, -5, -5, -5]^\top$. As in Section~\ref{sec:3d_experiment}, these values have been chosen based on our observations in simulations without the human of what weight combinations lead to the learning of successful trajectory distributions.

In~(\ref{eq:reward_features_6D_experiment}),  $d_{\text{start}}$ is the distance between the start pose of the trajectory and the desired start pose. The variable $d_{\text{end}}$ represents the distance between the end pose and the desired end pose. The feature $\mathcal{L}_{\text{wall}}$ prevents collisions with the obstacle (wall). The next two features prevent high velocities and accelerations, both Cartesian and rotational. Finally, the last feature prevents unnecessary rotations by punishing high Euler angles $\alpha$, $\beta$ and $\gamma$. The feature $\mathcal{L}_{\text{wall}}$ is computed according to (\ref{eq:L_obstacle}) with $d$ the minimum signed Euclidean distance between the pole and the wall. 

Ten users have been requested to perform two tasks as quickly as possible while avoiding collisions. Task 1 consisted of moving the pole from the start position to the end position through any of the two windows. Task 2 consisted of moving the pole from the start position to the end position through any of the two windows and bringing it back to the start position through the other window. The length of the pole is $2m$, the windows are squares with sides along $x$ and $z$ of $2m$, the wall is a square with sides along $x$ and $z$ of $100m$ and width (along $y$) of $3m$. The center of the wall is at $(x=5m, y=0m, z=-4m)$. The start position is at $(x=10m, y=-30m, z=-5m)$ and the end position at $(x=4m, y=20m, z=-5m)$.

\begin{figure}
\begin{center}
\includegraphics[width=\linewidth]{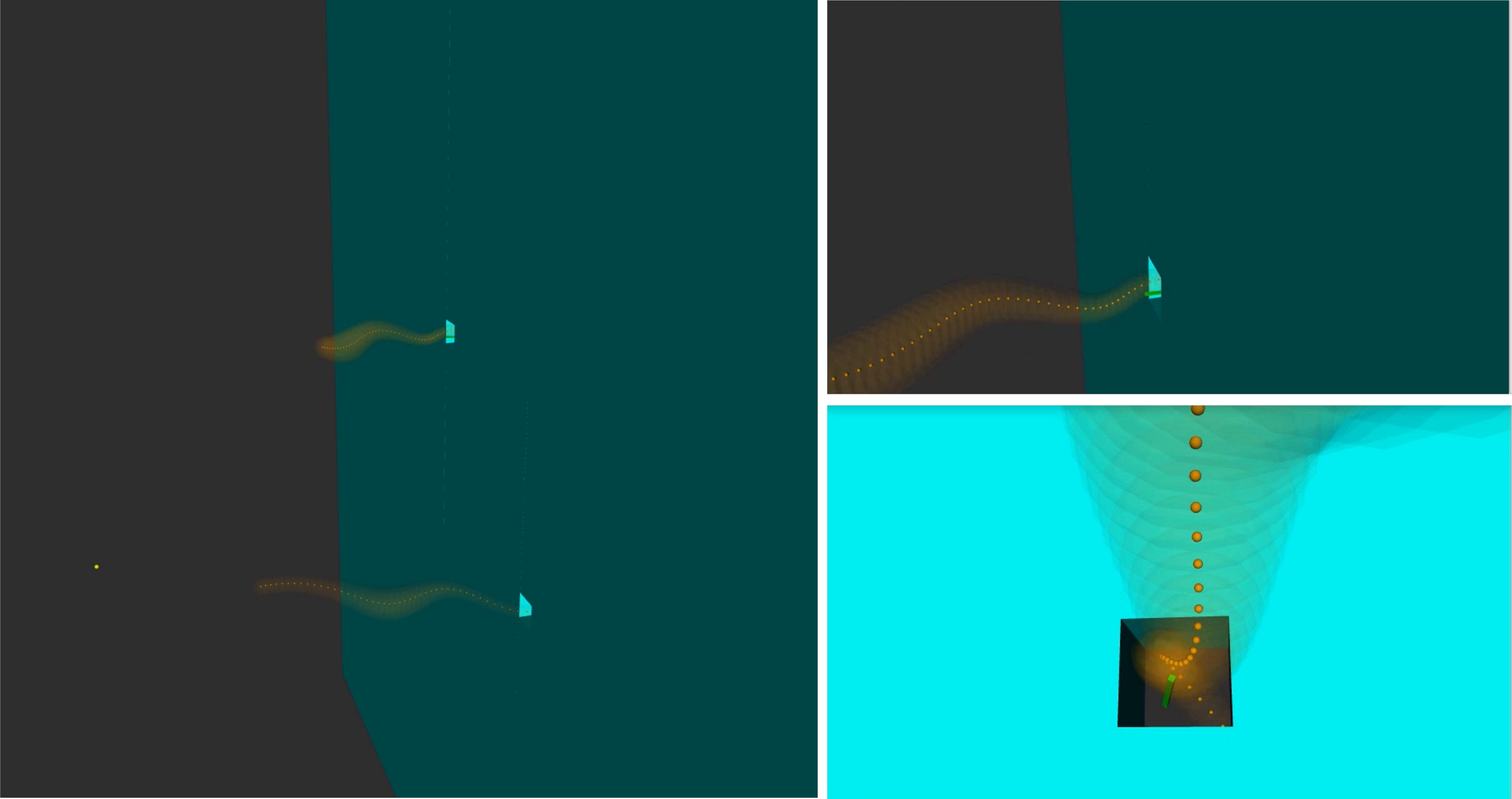}
\end{center}
\caption{Example of replanning in the experiment involving the teleoperation of a virtual pole through a haptic device. The little orange spheres represent positions along the mean of distributions of trajectories (ProMPs). The larger transparent orange spheres represent the variances in $x$, $y$ and $z$ corresponding to the positions depicted by the small spheres. Our framework learns ProMPs to help the user move the pole through the windows to a target position on the other side of the wall. When the user escapes the attraction of one of the ProMPs, another ProMP is learned to take the pole from its current position to the target position. The three images represent the same moment from three different perspectives.}\label{fig:6D_10}
\end{figure}

Task 1 has been performed by the users with two control modes: with force feedback or without force feedback. Task 2 has been performed with three control modes: with force feedback and replanning, with force feedback and without replanning, without force feedback. The order of the control modes experienced by each user has been determined at random. When force feedback was active, our system would compute a guiding ProMP. The initial ProMP resembles a tube starting at the start position and extending through one of the two windows until the end position. By applying a certain amount of force to the handle of the haptic device, users could escape the guidance of the initially planned ProMP. With replanning enabled, a new ProMP from the pole's current pose to the end pose would be planned once the user had escaped all the previously planned ProMPs. Without replanning, no new ProMP would be planned. Fig.~\ref{fig:6D_10} shows an example of replanning enabled.

\begin{wrapfigure}{R}{0.5\textwidth}
\centering
\subfloat[Teleoperation mode versus number of collisions]{%
\resizebox*{0.47\textwidth}{!}{\includegraphics{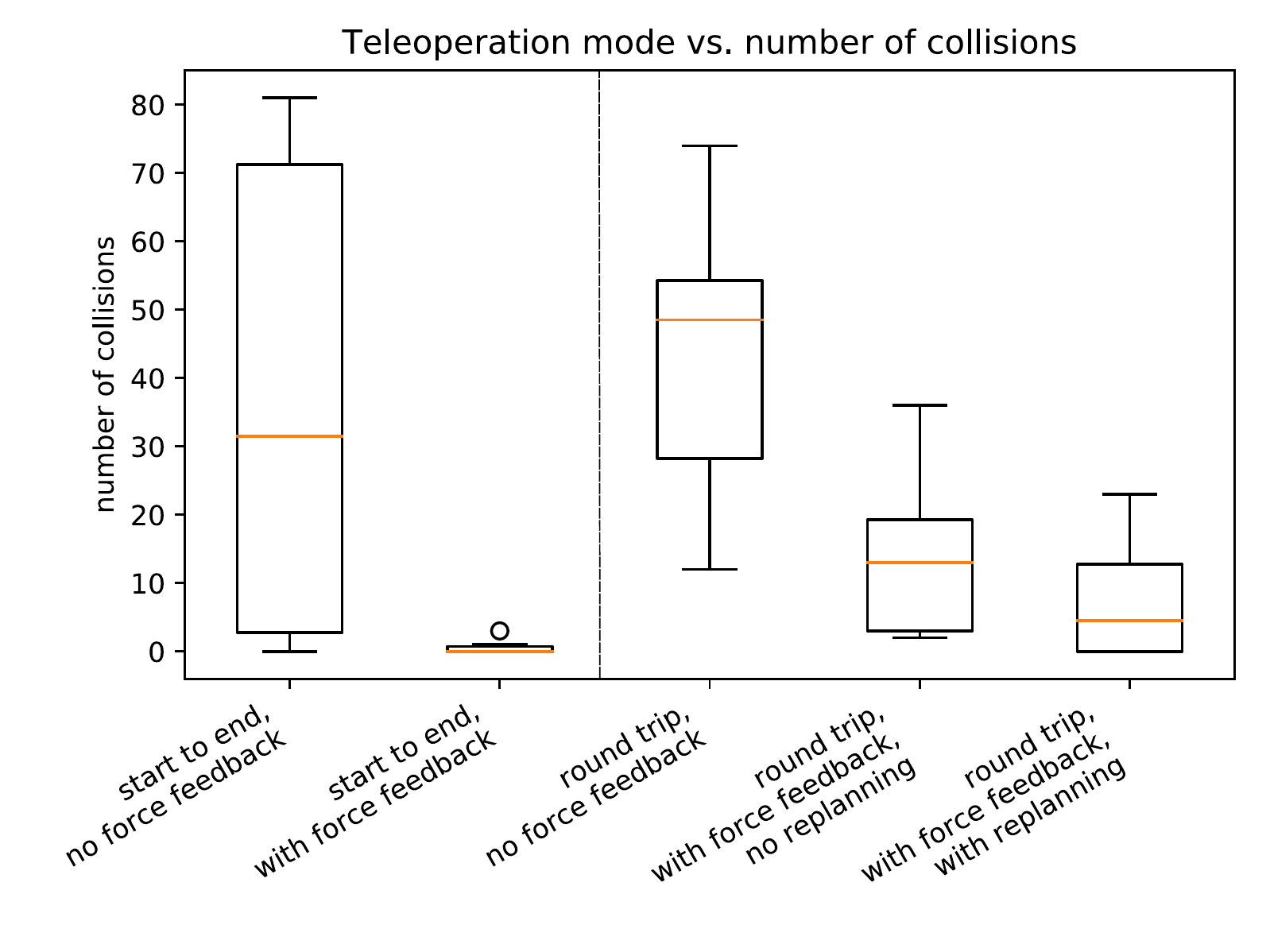}}}\vspace{5pt}
\subfloat[Teleoperation mode versus time to complete the task]{%
\resizebox*{0.47\textwidth}{!}{\includegraphics{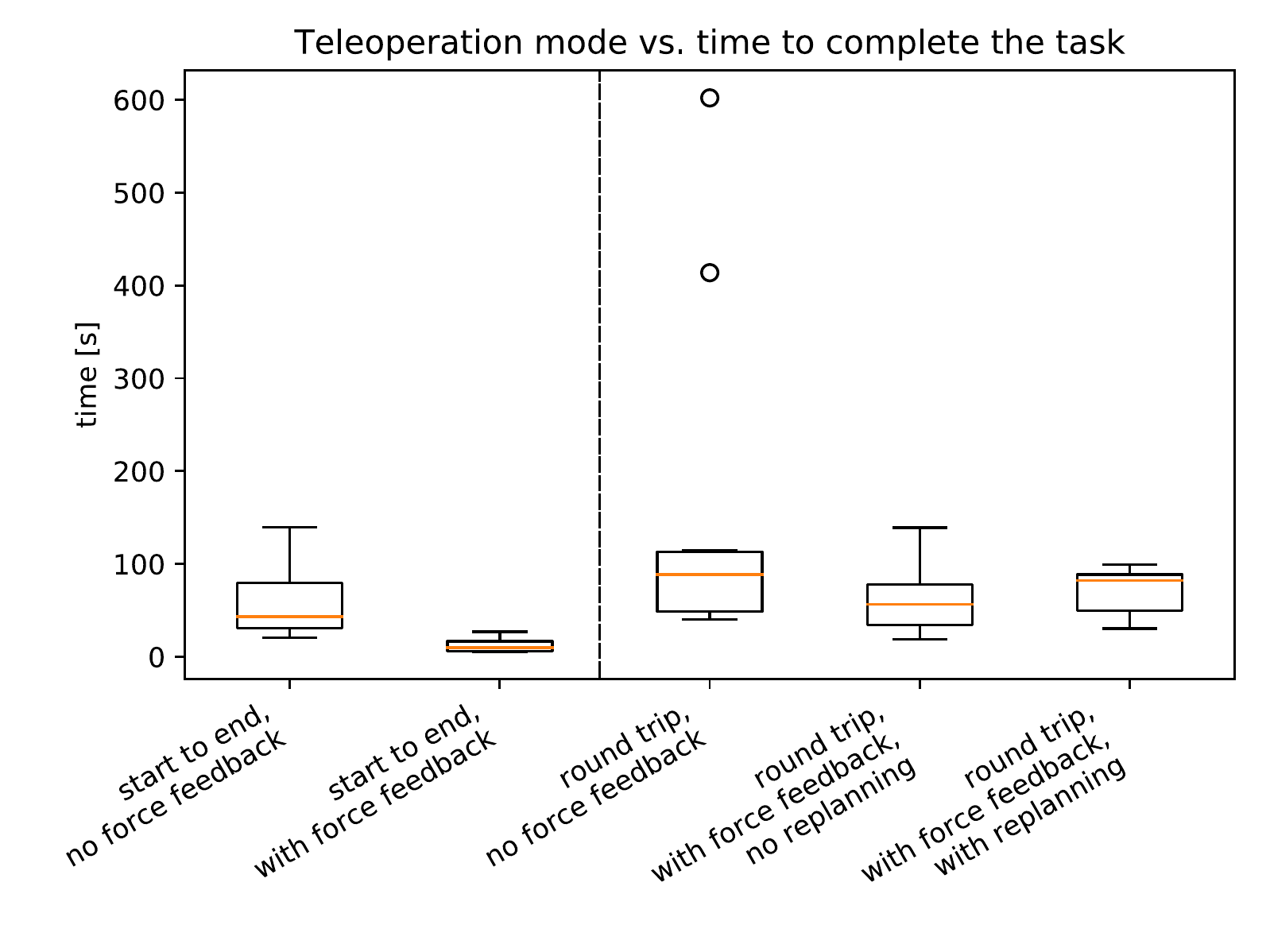}}}\vspace{5pt}
\subfloat[Teleoperation mode versus feeling of control]{%
\resizebox*{0.47\textwidth}{!}{\includegraphics{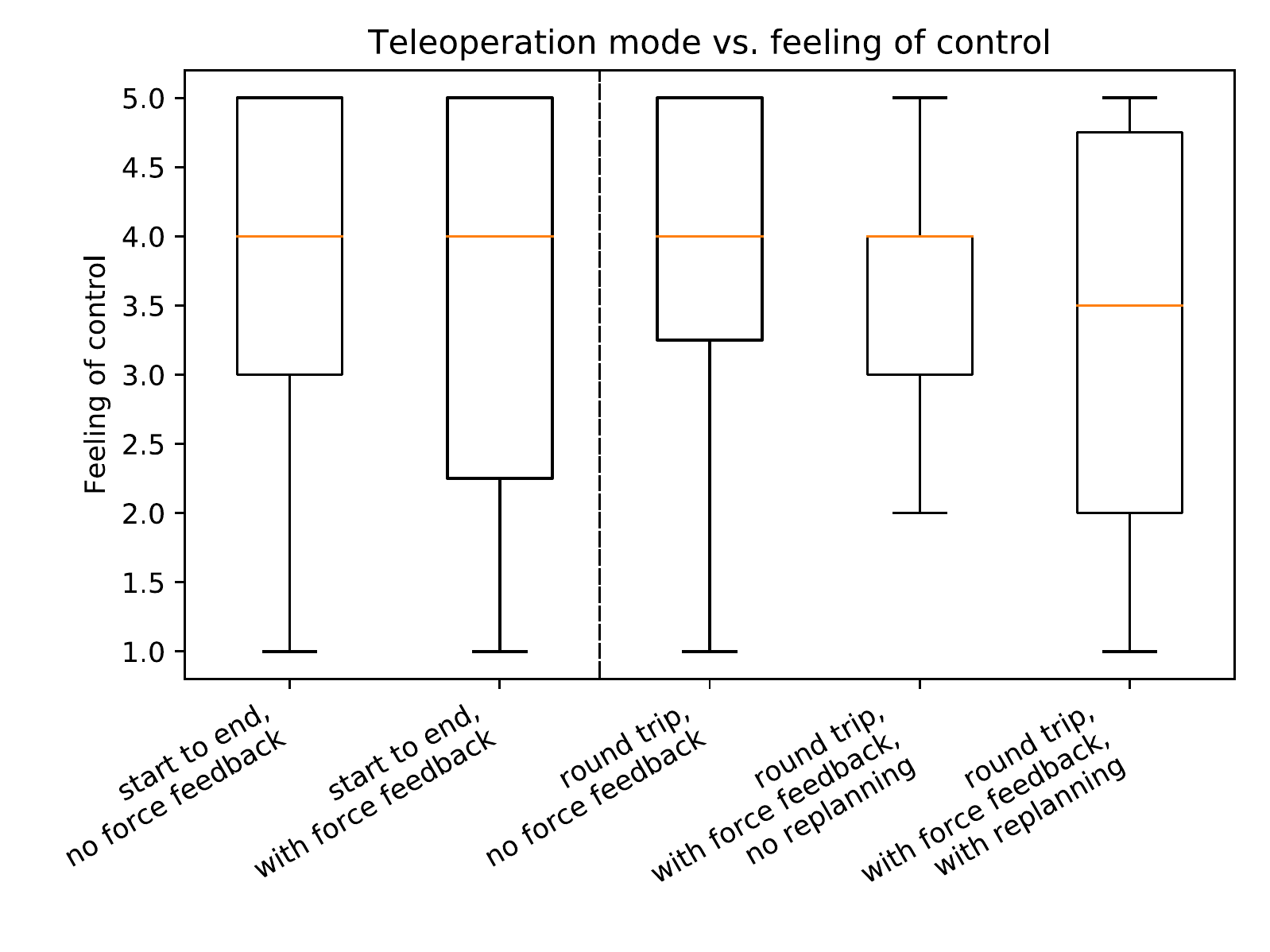}}}
\caption{Box plots representing the data acquired in our user studies on the control of the 6D pose of a virtual pole by manipulating a haptic device.}
\vspace{-3em}
\label{fig:user_studies}
\end{wrapfigure}

In summary, each user performed Task 1 (start to end) two times (with and without force feedback) and Task 2 (round trip) three times (with force feedback and replanning, with force feedback and no replanning, without force feedback). When executing Task 2 with force feedback and no replanning, the users have been asked to skip the ProMP planned from start to end position, pass through one of the windows without force feedback and use the originally planned ProMP as a guide only when returning to the start position. The number of collisions with the wall in each trial has been recorded as well as the time users took to complete each task. Task 1 was considered completed once the distance between the center of the pole and the end position (marked by the blue sphere) was under $4m$. Task 2 was completed once, after reaching the position marked by the blue sphere, the distance between the center of the pole and the start position (marked by the yellow sphere) was under $4m$. Moreover, after all trials, users have been asked ``On a scale from 1 to 5, how much did you feel in control of the haptic device? 1 means completely not in control. 5 means completely in control.'' Fig.~\ref{fig:user_studies} shows the results.

Nonparametric ANOVA Kruskal-Wallis hypothesis tests with a significance level $\alpha = 0.05$ have been conducted. For the evaluations involving more than two groups, post-hoc Conover's tests have been used to indicate which groups were significantly different.

There was a significant difference between ``with force feedback'' and ``no force feedback'' with respect to the number of collisions ($p = 0.0137$) and time to complete the task ($p = 0.0002$) in Task 1. The difference between `with force feedback'' and ``no force feedback'' with respect to the feeling of control reported by the users was not significant ($p = 0.9060$) in Task 1.

Kruskal-Wallis test has indicated a significant difference ($p = 0.0008$) between the three teleoperation modes in Task 2 with respect to the number of collisions. Conover's test has indicated that the differences were significant between ``no force feedback'' and ``with force feedback, no replanning'' ($p = 0.0035$) as well as between ``no force feedback'' and ``with force feedback, with replanning'' ($p = 0.0001$). Kruskal-Wallis test has indicated no significant difference ($p = 0.2050$) between the teleoperation modes in Task 2 with respect to the time users took to complete the task. There was also no significant difference ($p = 0.7563$) between the three teleoperation modes in Task 2 with respect to the feeling of control reported by the users.

Based on the box plots and hypothesis tests, we conclude that our framework helps users avoid collisions in both tasks. Furthermore, our framework helps users complete Task 1 faster than without assistance. Although the hypothesis tests did not indicate a significant difference between the teleoperation modes in Task 2 with respect to the time users took to complete the task, Fig.~\ref{fig:user_studies}(b) shows two outliers for the teleoperation mode without force feedback. The modes with force feedback did not present outliers. Therefore, our framework may help users to not require a very large amount of time to complete Task 2. Moreover, the box plots and hypothesis tests indicate that our framework helps users avoid collisions in Task 2 without compromising the time users take to complete the task.

The hypothesis tests indicate that the difference between the medians of the feeling of control reported by the users was not significant in any of the tasks. We conclude that our framework does not affect much the feeling of control of the users over the haptic device. This result is in accordance with the intended purposes of our framework, since it is supposed to assist users while giving them the freedom to diverge from the guidance of the haptic device.

%% file: conclusion.tex
This paper introduced a new approach to assisted teleoperation. A GMM of trajectories is learned through variational inference with the VIPS algorithm. VIPS learns a GMM that maximizes an episodic reward function in expectation while maintaining high entropy. Each mixture component is a ProMP. The gradient of the logarithm of the marginal distribution in end-effector space plus a damping term is equal to the wrench applied by the haptic device on the user. Moreover, Bayesian belief update is used to infer the plan and the phase intended by the user. The beliefs about the plan and the phase correspond to the weights of the mixture components in end-effector space. Therefore, these beliefs influence the wrench applied by the haptic device on the user.

Our approach can guide the user around obstacles when classical potential fields methods could get stuck in local minima. It can learn multimodal solutions to a task and adapt online to changes in the environment or in the intention of the user, as it has been demonstrated in our experiments. User studies have demonstrated that our framework can help users perform teleoperation tasks faster and avoiding collisions with obstacles.

A promising line of research for future work would be to combine the proposed framework with inverse reinforcement learning to avoid specifying the weights of the features of the reward function manually. In addition, user studies involving the teleoperation of a real robot arm would be interesting to evaluate the efficacy of our approach in more realistic scenarios. In this case, our system should also give the user haptic cues to avoid singularities and joint limits.